\documentclass[runningheads]{llncs}

\usepackage[T1]{fontenc}
\usepackage[colorlinks=true,
            linkcolor=blue,
            citecolor=blue,
            urlcolor=blue]{hyperref}
\usepackage{tikz}
\usetikzlibrary{arrows.meta, positioning, fit, backgrounds, calc}
\definecolor{cData}{RGB}{30,90,168}
\definecolor{cMethod}{RGB}{201,120,20}
\definecolor{cModel}{RGB}{34,139,64}
\usepackage{url}
\definecolor{cResult}{RGB}{124,58,160}
\usepackage{amssymb} 
\usepackage{svg}
\usepackage{pgfplots}
\pgfplotsset{compat=1.18}
\usepgfplotslibrary{groupplots}
\usepackage{graphicx}
\usepackage{amsmath}
\usepackage{xcolor}
\usepackage{booktabs}
\usepackage{orcidlink}
\usepackage{verbatim}
\usepackage{microtype}
\providecommand{\orcidlink}[1]{\href{https://orcid.org/#1}{\orcidlogo}}

\begin{document}

\title{Sensitivity Analysis of Decision Rules in an Agent-Based Agro-Ecological Model}
\titlerunning{Sensitivity Analysis of Decision Rules for Agro-Ecological ABM}
\author{
Benjamin Pianet\inst{1,2}\orcidlink{0009-0001-5708-9312} \and
Baptiste Lesquoy\inst{1,2,3}\orcidlink{0009-0000-6906-9418} \and
Paul Saves\inst{1,2,3}\orcidlink{0000-0001-5889-2302} \and
Benoit Gaudou\inst{3}\thanks{Corresponding author.}\orcidlink{0000-0002-9005-3004} \and
Jean Villerd\inst{4}\orcidlink{0000-0001-7156-7512} \and
Nicolas Verstaevel\inst{3}\orcidlink{0000-0002-7879-6681} \and
Olivier Therond\inst{4}\orcidlink{0000-0003-0191-3509}}

\authorrunning{B. Pianet et al.}

\institute{
$\ ^1$ACROSS IJL, IRD, Thuyloi University,  Hanoi, Vietnam \\
$\ ^2$IRD/UMMISCO, Sorbonne Universit\'e,  Hanoi, Vietnam \\
$\ ^3$Université Toulouse Capitole, IRIT, Toulouse, France \\
$\ ^4$ Laboratoire Agronomie et Environnement, INRAE, Colmar, France \\
\email{benoit.gaudou@ut-capitole.fr}
}
\maketitle

    \begin{abstract}
 Agent-based models are well-suited for modelling agro-ecological systems for decision-making, yet turning their simulations into actionable insight requires identifying which decisions actually drive the behaviours.
In such a context, sensitivity analysis stumbles upon a structural feature of many complex systems: the parameters are not a flat vector but a decision tree in which many parameters appear only when a higher-level choice activates them. 
Classical global sensitivity analysis presupposes independent inputs over a fixed Cartesian product and is ill-posed on such a branched space and biased by construction.
We instead treat the structure of the decisions as a first-class modelling object.
Using advanced surrogate modelling techniques, we encode the crop-management itinerary of the high-resolution agro-ecological modelling platform MAELIA as a hierarchical design space featuring both categorical and numerical variables. 
We propose a novel dedicated method to analyse this kind of hierarchical space, a random-forest-supervised HSIC-ANOVA, which ranks main effects and interactions natively on the tree while placing rarely and always active decisions on an equal footing. 
We show how, when applied to simulation outputs such as nitrogen leaching, soil organic carbon variation, and yield predictions, it can deliver an interpretable, decision-oriented reading.

\end{abstract}

\section{Introduction}

Socio-agro-ecological systems couple human decisions with biophysical processes such as crop growth, water flow, and nutrient cycling. Agent-Based Models (ABMs) have become a reference tool for representing these couplings because they relax the strong analytical assumptions of aggregate models and make space, heterogeneity, stochasticity and individual decision-making explicit~\cite{tenbroeke2016which,ligmann2014socioecological}. The MAELIA (Modelling of socio-Agro-Ecological system for Landscape Integrated Assessment) platform is a representative example: built on the GAMA simulation platform~\cite{taillandier2019gama}, it simulates the interaction between agricultural activities, hydrology and water-resource governance at a daily time step and from the plot to the territory spatial scale, and it produces a broad range of socio-economic and agro-environmental indicators~\cite{gaudou2014maelia,therond2014maelia}. Still, this descriptive variety makes the model difficult to analyse, since any simulated output emerges from the coupling of its components, computed jointly at each time step, each feeding the others, rather than from a single parameter taken in isolation. Attributing an output to the inputs is further complicated by the shape of the input space itself. In such applications, parameters form a \emph{decision tree} rather than a flat vector: \emph{conditionally active} sub-parameters have meaning only when decreed by a higher-level \emph{meta} variable, so the very existence of a parameter depends on the value taken by another. Applying a flat sampling strategy to this branched space generates invalid architectural combinations, typically handled by imputing default values to inactive decreed variables~\cite{sportisse2020imputation}. Therefore, a structured hierarchical treatment of the parameter space is necessary~\cite{saves2026hierarchical} as formalised in~\cite{audet2023general}. Moreover, the Surrogate Modelling Toolbox (SMT) and adsg-core libraries provide open-source software to deal with them practically~\cite{saves2024smt}.
To understand such complex models and quantify how much of the output variability can be assigned to the inputs, practitioners generally perform Global Sensitivity Analysis (GSA) ~\cite{razavi2021future}. While a substantial literature adapts these tools to ABMs specifically~\cite{tenbroeke2016which,thiele2014cookbook,borgonovo2022protocol}, two obstacles remain in GSA for complex systems. First, variance-based indices require many model evaluations, which are prohibitive for a time-consuming simulator (which is the case most of the time with ABM). Second, and more fundamentally, classical GSA assumes \textit{i.i.d.}\ (independent and identically distributed) inputs, which a hierarchical space violates by construction~\cite{saltelli2008global}. Furthermore, imputed values mask the true influence of the upstream decisions that actually control a branch, breaking the variance decomposition. To overcome this, a complementary line of work makes the exploration itself systematic: a recent multi-stage workflow for stochastic ABMs first screens the space and then trains a surrogate on which it computes sensitivity indices and explanations~\cite{saves2026mabs}. 

Whether such an approach transfers to a complex, real agro-ecological simulator whose parameter space is genuinely hierarchical remains an open challenge.
This paper addresses this challenge. We model the structure of agricultural decisions and analyse the sensitivity of MAELIA within it. Our contributions are threefold:
    (i) \textbf{An application to a complex model:} We apply the analysis to MAELIA, a high-resolution agro-ecological agent-based modelling platform whose daily-step coupling of management, soil and climate is markedly more complex than the toy or single-purpose models that dominate the ABM sensitivity-analysis literature;
  (ii) \textbf{A hierarchical, structured parameter space for SA:} We encode MAELIA's crop-management decision as a design space of meta and decreed variables (soil-preparation and fertilisation switches activating their dependent counts, depths, doses and dates),
  spanning $14$ potential parameters, so that sampling and sensitivity analysis respect the conditional activation of variables;
  (iii) \textbf{A structure-aware analysis pipeline:} We propose a two-stage pipeline over the hierarchical space that couples random-forest supervision with an HSIC-ANOVA decomposition. The random forest identifies the influential variables, and the subsequent decomposition quantifies how strongly the outputs depend on them, handling categorical and numerical inputs alike.

The remainder of this paper is as follows.
Section~\ref{sec:related} reviews sensitivity analysis methods for agent-based models, hierarchical and mixed-variable design spaces, and surrogate-assisted exploration. Section~\ref{sec:background} presents MAELIA and its parameterisation through date-dose files. Section~\ref{sec:method} details the proposed analysis pipeline. Section~\ref{sec:results} reports the experimental setup and the results. Section~\ref{sec:conclusion} concludes and outlines perspectives.

\section{Related Work}
\label{sec:related}
\begin{table}[bt!]
\centering
\caption{Attributes of the date-dose formalism. The depth, dose, and subtype apply only to the listed operations and are left empty otherwise.}
\label{tab:format}
\resizebox{\linewidth}{2.5cm}{
\begin{tabular}{@{} p{2cm} p{3.8cm} p{6.5cm} c @{}}
\toprule
\textbf{Attribute} & \textbf{Applies to} & \textbf{Admissible values} & \textbf{Unit} \\
\midrule
Plot      & All operations & Any plot in the simulation & --- \\
Species   & All operations & Any species in the simulation & --- \\
Date      & All operations & Any calendar date (DD/MM/YYYY) & --- \\
Operation & All operations & Sowing, tillage, hoeing, fertilisation, plant protection, irrigation, mowing, grazing, harvest & --- \\
\midrule
Depth     & Sowing, tillage, hoeing & $\geq 0$ (or empty for no effect on water reserve) & cm \\
          & Mowing                  & $\geq 3$ & cm \\
\midrule
Dose      & Fertilisation (mineral) & $\geq 0$ (N required; P, K optional) & kg/ha \\
          & Fertilisation (organic) & $\geq 0$ & kg/ha \\
          & Irrigation              & $\geq 0$ & mm \\
\midrule
Subtype   & Fertilisation           & Any fertiliser in the simulation & --- \\
          & Other operations        & Any free label (or empty) & --- \\
\bottomrule
\end{tabular}
}
\end{table}
\textbf{Sensitivity Analysis of Agent-Based Models.}
Sensitivity analysis attributes model output variability to its inputs. Methods range from low-cost Morris screening for initial factor ranking~\cite{morris1991factorial} to variance-based decomposition. For instance, Sobol' indices capture main and interaction effects~\cite{sobol2001global,saltelli2002best}, while Shapley effects allocate variance even with dependent inputs~\cite{shapley1953value,song2016shapley}. Applying GSA to ABMs requires addressing intrinsic stochasticity, output selection, and replication limits, which a dedicated literature addresses via practical guidelines and tailored protocols~\cite{thiele2014cookbook,tenbroeke2016which,borgonovo2022protocol}. Although variance-based GSA successfully enhances the analytical and policy relevance of agro-ecological ABMs~\cite{ligmann2014socioecological,baniecki2024robustness}, these approaches are inadequate for models governed by optional, hierarchical operations.

\noindent \textbf{Hierarchical and Mixed-Variable Design Spaces.}
Most GSAs assume \textit{i.i.d.} inputs ranging over a Cartesian product. Agricultural itineraries break this assumption: variables are \emph{mixed} (continuous dates and doses, integer counts, categorical operations, and fertiliser types) and \emph{hierarchical}, \textit{i.e.}\ some variables are active only under specific values of others. A unified framework describes such spaces through meta-variables (which decide which sub-problem is instantiated) and decreed variables (active only when their parent decision enables them), together with the graph-structured distances needed to model them~\cite{audet2023general,saves2026hierarchical,hallehannan2025distance}. 

\noindent  \textbf{Surrogate-Assisted, Multi-Stage Exploration.}
Because ABM simulations are costly, exhaustive GSA is rarely affordable directly~\cite{klugl2008validation}. Training surrogates such as polynomial chaos, random forest, or Gaussian process to compute sensitivity indices is efficient in that context~\cite{rasmussen2006gaussian,angione2022surrogate,geurts2006extremely}.
Building on this, a recent two-phase workflow screens the space using space-filling designs, linear analysis, and regression trees to rank factors, size replications, and locate tipping points, before training a surrogate for indices and explanations~\cite{saves2026mabs}. We adapt this approach to a more complex model, filling a critical gap by representing the agricultural parameter space as a genuinely hierarchical directed acyclic graph.

\section{The MAELIA platform}
\label{sec:background}

\subsection{Main Features and Capabilities}
The MAELIA (Modelling of socio-Agro-Ecological system for Landscape Integrated Assessment) platform is a spatially explicit, agent-based platform for the integrated assessment of socio-agro-ecological systems~\cite{gaudou2014maelia,therond2014maelia,misslin2021challenges}. 
The latter covers a family of processes such as agronomic, hydrological, or socio-economic dynamics, allowing each configuration to be specifically tailored for the emulation of any given territory and research question.
At a daily time step and at the individual field and farmer levels, it brings together the main processes that shape an agricultural territory. Driven by daily weather, it simulates the growth of crops, the water, carbon, and nitrogen cycles in the soil, the dynamics of water resources, and the farmers' activities on the land, including withdrawals for irrigation. 
Each simulated day, these processes converge into a specific situation that each farmer agent reads and acts upon. For instance, considering past and forecasted weather along with soil water content, a farmer may hold off sowing, while also navigating structural constraints such as regulatory drought decrees~\cite{therond2014maelia}. Whichever technical operations (tillage, sowing, fertilisation...) are ultimately performed on the crops or the soil, they in turn modify plant growth and resource uptake, altering the very state from which the next day's decisions will be made. MAELIA is built to capture exactly these kinds of interactions, scaling seamlessly from a small cluster of farms to large territories comprising 1,150 agricultural holdings across 40,000 hectares~\cite{tribouillois2022integrated}.

MAELIA is a modular platform assembled from many sub-models and modelling options that can be activated according to the question at hand~\cite{misslin2021challenges}.
Demonstrating its versatility, MAELIA can model the quantitative management of water resources during low-flow periods~\cite{mazzega2014critical}, while other configurations have addressed territorial water management~\cite{martin2016use,tribouillois2022introducing,therond2014maelia}, crop-livestock systems~\cite{catarino2021fostering}, organic waste management~\cite{misslin2022integrated}, and agroecological transitions like cover crop adoption~\cite{tribouillois2022integrated}.
MAELIA has been applied to many territories in France and Germany, and the current work aims to scale it to all of France~\cite{javourez2025maelia}.

\subsection{MAELIA architecture}
\label{ssec:architecture}
Spatially, a MAELIA simulation relies on a multi-layered topology in which the territory is partitioned along distinct, intersecting boundaries. By computing processes across these overlapping layers, the model aggregates them into a coherent territorial dynamic. Partitions include: (i) meteorological zones matching the ARPEGE (Météo-France) grid; (ii) administrative boundaries for communes; (iii) hydrological watersheds; and (iv) agricultural fields governed by the European land parcel identification system. Weather information is read from external files (attached to weather zones); rainfall drives water balance and drainage, while temperature and radiation drive crop growth. 
The soil is described by its type and refined into layers of possibly unequal thickness, with general properties such as bulk density, gravel/clay contents, water-holding capacity or maximal depth.

The simulation biophysical representation rests on the crop model 
together with nitrogen and carbon 
modules~\cite{catarino2021fostering,misslin2021challenges}. Henceforth, this study considers the following outputs: (i) \textbf{Nitrogen leaching (\(\text{N}_{\text{lixi}}\))}: the deep drainage flux leaving the soil profile when water inputs exceed the field capacity of the deep horizon, carrying dissolved mineral nitrogen with it;
   (ii) \textbf{Variation of soil organic carbon (\(\Delta \text{C}_{\text{org}}\))}: a daily balance between the humification of crop residues and the mineralisation of active and stable soil organic matter, modulated by temperature and moisture;
   (iii) \textbf{Yield (\(\text{Yield}\))}: simulated from a potential optimal yield, penalised daily by water stress, the ratio of actual to maximal transpiration, and by nitrogen stress through the nitrogen nutrition index.
   
Farmers are the agents at the centre of this work, operating at two distinct decision-making scales~\cite{gaudou2014maelia}. Annually, the farmer draws up a cropping plan, allocating crop rotations across the plots they manage. The crop grown on a given plot can either follow a predefined sequence or be dynamically selected by the agent. The latter mode, which yields richer emergent behaviour, relies on a multi-criteria decision process grounded in Dempster-Shafer belief theory within a belief-desire-intention architecture~\cite{taillandier2012bdi}. Subsequently, at the daily scale, the farmer manages the crop through technical operations: soil tillage, sowing, fertilisation, irrigation, hoeing, and harvesting. The dated sequence of operations applied to a crop is called a \emph{crop management itinerary} and constitutes the object of our study.
These operations are governed by \emph{decision rules} elicited from domain experts and supplied as input to the model. This approach ensures that the resulting simulated behaviours remain realistic and reusable across different territories, crops, and weather conditions~\cite{bergez2012tools}. Specifically, each technical operation is confined to a temporal window and enabled only when specific conditions of the crop, the soil, and the weather are satisfied. Furthermore, whether an operation is actually performed, and to what extent, is constrained by the workforce available on the farm~\cite{gaudou2014maelia}. 
Weather forecasts also play a role in this process, influencing the decisions of certain agents based on the reliability the agents attribute to them. 
These decisions couple the sub-models mainly through the state of the soil and its plot: a water-balance sub-model updates soil moisture, the crop sub-model reads it to advance growth, the farmer reads both to decide whether an operation can take place, and the operation rewrites the plot state for the next day. This indirection through the soil keeps the sub-models interchangeable, so a component can be deactivated without disturbing the rest~\cite{gaudou2014maelia}.

At this second scale, the agents' decisions meet the biophysical mechanisms, and each output is sensitive to them: tillage reworks the soil and the mineralisation of the residues it buries, fertilisation sets the nitrogen made available to the crop, and the sowing and harvest dates bound the growing window. Because the rules tie each operation to the evolving state of the crop environment, the model keeps farming decisions adaptive yet interpretable, any operation being traceable to the rule that triggered it~\cite{drogoul1994multi}.
Both scales of decision ultimately rest on rules, and the values of its parameters specify every rule. Those parameters, how they are set, and the alternatives we substitute for them, are addressed in what follows.

\subsection{Model Parameters}
\label{ssec:datedose}

\noindent {\textbf{A Controllable input space for crop management itineraries.}
\label{ssec:controllable-input-space}
This article focuses on the agricultural model and, in particular, on the technical operations that compose the farmer agents' technical itineraries. This study examines the effect of these operational choices on three outputs  ($N_\text{lixi}, \Delta C_\text{org},$ Yield).
The natural input space is that of the crop management itineraries themselves, but it is not directly accessible in MAELIA: technical operations are triggered depending on farmers' decision rules, and an itinerary is thus fully known only at the end of a simulation.
Sampling the rule-related parameters explores the itinerary space redundantly, incompletely, and with little control.
We therefore added to MAELIA a new way of specifying technical operations, one that describes the itinerary directly and bypasses the decision rules altogether. Such a description, as a list of dated operations with their quantities, is a common formalism, used as input or output by many agricultural models~\cite{javourez2025maelia} that we adapted to the specificity of MAELIA. We introduce a new type of farmer agent driven by this \emph{date-dose} formalism. This agent is functionally equivalent to the rule-based one: an itinerary produced by a rule-based simulation, extracted and replayed through the date-dose agent, yields identical outputs and intermediary variables. What changes is only how the itinerary is obtained: stated rather than inferred.

\noindent {\textbf{The date-dose formalism.}}
\label{ssec:datedose-formalism}
In the date-dose formalism, an itinerary is a list of technical operations, each described by seven attributes. Two locate the operation: the plot it is applied to and the cultivated species. 
One gives its type, drawn from the operations that MAELIA can natively handle: \emph{sowing}, \emph{tillage}, \emph{hoeing}, \emph{fertilisation}, \emph{plant protection}, \emph{irrigation}, \emph{mowing}, \emph{grazing}, and \emph{harvest}. One gives its execution date. And, the last three apply only to the operations for which they are meaningful: one encodes a working depth or cutting height, another a dose, whose unit depends on the operation and, the last one a subtype, which for a fertilisation names the fertiliser applied and for any other operation is a free label with no effect on the run, used only to retrieve the tagged operations in the model's outputs. Table~\ref{tab:format} specifies each attribute and its admissible values.

\noindent {\textbf{The crop-management decision as a hierarchy.}}
\label{ssec:hierarchy-detail}
Of the operations MAELIA recognises, the study explores only sowing, soil preparation, fertilisation and harvest, and it realises soil preparation by tillage alone, leaving the other operation types such as hoeing, plant protection, irrigation, mowing and grazing outside the design space.
The attributes introduced in this section are structured as a hierarchy of conditional decisions spanning multiple levels of the technical itinerary. A first meta variable, the number of tillage passes (\nolinkurl{Number\_Of\_Preparations} $\in \{0,1,2\}$), decides whether soil preparation occurs at all and, when it does, how many passes are performed. Each preparation event is then described by an explicit working depth, namely \nolinkurl{First\_Tillage\_Depth} for the first operation and, when a second operation is active, \nolinkurl{Second\_Tillage\_Depth}. The timing of soil preparation is expressed through a temporal offset before sowing (\nolinkurl{Preparation\_To\_Sowing\_Delay}), which keeps the preparation operation agronomically ordered with respect to sowing.
Sowing is characterised by its campaign date (\nolinkurl{Sowing_Date}) and by a sowing depth (\nolinkurl{Sowing_Depth}), the latter included because sowing also works the upper soil layer and thus bears on carbon and nitrogen redistribution.
Similarly, the fertilisation block follows a conditional structure~\cite{saves2026hierarchical}. The number of fertilisation events is controlled by a second meta variable, \nolinkurl{Number\_Of\_Fertilisations} $\in \{0,1,2,3\}$. If no fertilisation is selected, no fertilisation date or dose is active. If one, two, or three events are selected, the corresponding dates (\nolinkurl{First\_Fertilisation\_Date}, \nolinkurl{Second\_Fertilisation\_Date}, \nolinkurl{Third\_Fertilisation\_Date}) and doses (\nolinkurl{First\_Fertilisation\_Dose}, \nolinkurl{Second\_Fertilisation\_Dose}, \nolinkurl{Third\_Fertilisation\_Dose}) become active. Fertilisation is represented through a single fertiliser reference, so that the explored variability focuses on the timing and amount of nitrogen supplied rather than on product identity.

To ensure chronological consistency, operational dates are constrained relative to one another. Sowing, harvest, and fertilisation dates are expressed in campaign days from August~$1^\text{st}$ (day $1$). The design enforces agronomic temporal order: soil preparation must occur before sowing, fertilisation events must follow their activated chronological sequence, and harvest must occur after sowing~\cite{saves2026hierarchical}. The full decision structure is formalised as the design space graph of Fig.~\ref{fig:space}, comprising the $14$ potential variables---spanning meta, discrete, and continuous types---intertwined through conditional activation and value constraints~\cite{bussemaker2025system}.
Specifying the itinerary directly gives full control over the operations, but the model no longer guarantees that the result makes agronomic sense: a date-dose file can schedule a harvest before a sowing. Turning this formalism into an input space for the sensitivity analysis, one restricted to agronomically admissible itineraries and given a well-defined structure, is the subject of the next section.

\section{Sensitivity Analysis over a Hierarchical Space}
\label{sec:method}

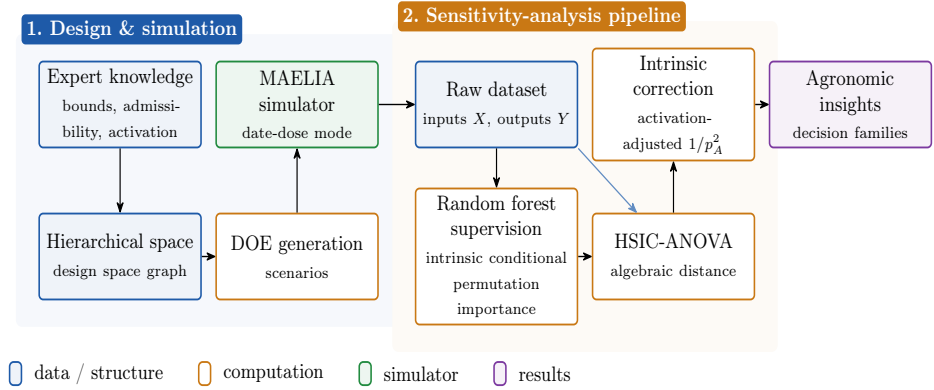
\begin{figure}[tbp!]
\centering
\resizebox{\textwidth}{5.15cm}{%
\begin{tikzpicture}[
  >={Stealth[round]},
  font=\footnotesize,
  box/.style={rectangle, thick, align=center, text width=26.7mm,
              minimum height=13mm, inner sep=3pt, rounded corners=2pt},
  data/.style   ={box, draw=cData,   fill=cData!7},
  method/.style ={box, draw=cMethod, fill=white},
  model/.style  ={box, draw=cModel,  fill=cModel!8},
  result/.style ={box, draw=cResult, fill=cResult!7},
  flow/.style   ={->, semithick},
  sub/.style    ={font=\footnotesize}
]

\node[data]  (ek)     at (0,0)      {Expert knowledge\\[1pt]{\scriptsize bounds, admissibility, activation}};
\node[model] (maelia) at (3.15,0)   {MAELIA simulator\\[1pt]{\scriptsize date-dose mode}};
\node[data]  (hs)     at (0,-2.3)   {Hierarchical space\\[1pt]{\scriptsize design space graph}};
\node[method](doe)    at (3.15,-2.3){DOE generation\\[1pt]{\scriptsize scenarios}};
\draw[flow] (ek)  -- (hs);
\draw[flow] (hs)  -- (doe);
\draw[flow] (doe) -- (maelia);

\node[data]   (rd)   at (6.7,0)     {Raw dataset\\[1pt]{\scriptsize inputs $X$, outputs $Y$}};
\node[method] (intr) at (9.85,0)    {Intrinsic correction\\[1pt]{\scriptsize activation-adjusted $1/p_A^{2}$}};
\node[method] (rf)   at (6.7,-2.3)  {Random forest supervision\\[1pt]{\scriptsize intrinsic conditional permutation importance}};
\node[method] (hsic) at (9.85,-2.3) {HSIC-ANOVA\\[1pt]{\scriptsize algebraic distance}};
\draw[flow] (rd)   -- (rf);
\draw[flow] (rf)   -- (hsic);
\draw[flow] (hsic) -- (intr);
\draw[flow, cData!70] (rd.south east) -- ([xshift=8mm]hsic.north west);

\node[result] (res) at (13.0,0) {Agronomic insights\\[1pt]{\scriptsize decision families}};
\draw[flow] (intr) -- (res);

\draw[flow] (maelia) -- (rd);

\begin{scope}[on background layer]
  \node[fill=cData!4,   rounded corners=3pt, inner sep=4mm,
        fit=(ek)(hs)(doe)(maelia)] (p1) {};
  \node[fill=cMethod!4, rounded corners=3pt, inner sep=4mm,
        fit=(rd)(rf)(hsic)(intr)] (p2) {};
\end{scope}
\node[anchor=north west, font=\footnotesize\bfseries, text=white,
      fill=cData, rounded corners=2pt, inner sep=2.5pt]
      at ([shift={(1mm,3mm)}]p1.north west) {1.\ Design \& simulation};
\node[anchor=north west, font=\footnotesize\bfseries, text=white,
      fill=cMethod, rounded corners=2pt, inner sep=2.5pt]
      at ([shift={(1mm,3mm)}]p2.north west) {2.\ Sensitivity-analysis pipeline};

\begin{scope}[shift={([yshift=-7mm]p1.south west)}]
  \node[data,   minimum height=3.5mm, text width=0pt, inner sep=3pt] (g1) {};
  \node[right=1mm of g1, sub] (h1) {data / structure};
  \node[method, minimum height=3.5mm, text width=0pt, inner sep=3pt, right=5mm of h1] (g2) {};
  \node[right=1mm of g2, sub] (h2) {computation};
  \node[model,  minimum height=3.5mm, text width=0pt, inner sep=3pt, right=5mm of h2] (g3) {};
  \node[right=1mm of g3, sub] (h3) {simulator};
  \node[result, minimum height=3.5mm, text width=0pt, inner sep=3pt, right=5mm of h3] (g4) {};
  \node[right=1mm of g4, sub] (h4) {results};
\end{scope}

\end{tikzpicture}}
\caption{Workflow of the data-driven hierarchical sensitivity-analysis pipeline. Expert knowledge fixes the bounds, admissibility constraints and conditional activation that define the hierarchical space. Scenarios are
simulated in MAELIA's date-dose mode. The resulting dataset feeds a random forest that \emph{supervises} the HSIC-ANOVA decomposition by setting its kernel scales, and a frequency correction yields the intrinsic indices.}
\label{fig:pipeline}
\end{figure}

The pipeline takes a table of MAELIA simulations, each row pairing a crop-management itinerary, \textit{i.e.}, a value for each of the 14 parameters, with the agro-environmental outputs it produces, and returns, for every output, a ranked list of the decisions and combinations of decisions that drive it. What sets it apart is that this ranking is computed \emph{on the design space graph}, without averaging away or imputing the variables that a higher-level choice leaves inactive.

Figure~\ref{fig:pipeline} summarises the procedure. Its upper panel covers design and simulation: expert knowledge fixes the bounds, the admissibility constraints and the conditional activation defining the hierarchical space, from which a design of experiments is drawn and run through MAELIA in date-dose mode. Its lower panel is our novel analysis, the random forest-supervised HSIC-ANOVA. It encodes the itinerary together with the pattern of which variables are active (Section~\ref{ssec:encoding}), lets a random forest learn which variables matter for each output and prune the rest to keep the analysis tractable (Section~\ref{ssec:rf}), ranks main effects and interactions with a kernel-based dependence measure that inactive branches cannot bias (Section~\ref{ssec:hsic}), and rebalances parameters active only part of the time against those always active (Section~\ref{ssec:intrinsic}). We then discuss its novelty (Section~\ref{ssec:novelty}).

\begin{figure}[tb!]
    \centering
    \resizebox{\linewidth}{7cm}{%
        \includegraphics{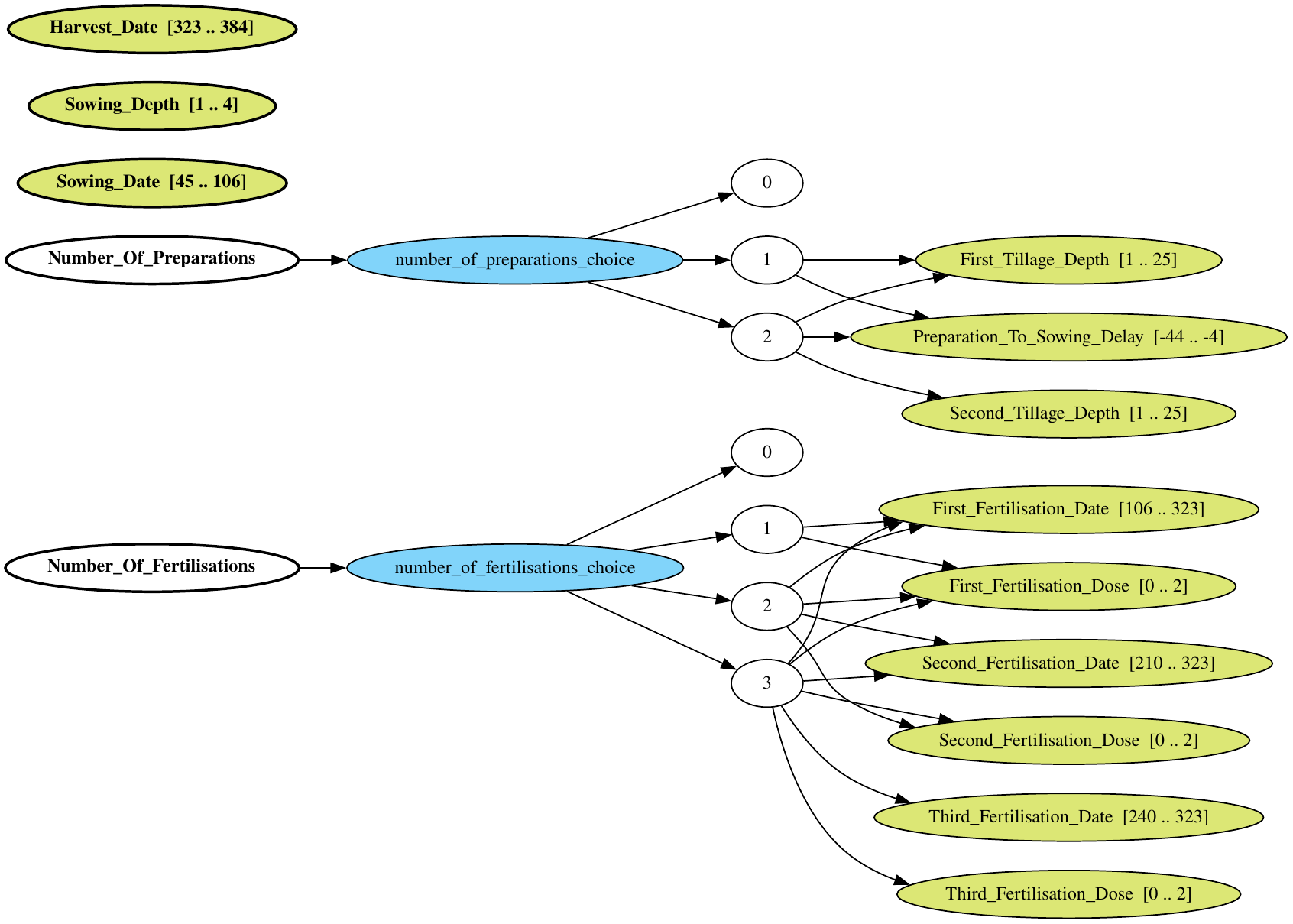}%
    }
    \caption{Design space graph of the MAELIA crop-management decisions.
    Each structural count is a meta variable feeding a choice node (blue),
    whose branches enumerate its admissible values; selecting one activates
    the decreed variables (yellow). The sowing date, the sowing depth and
    the harvest date depend on no choice and are always active.
    }
    \label{fig:space}
\end{figure}
\subsection{Encoding the Itinerary and its Activation Pattern}
\label{ssec:encoding}

Figure~\ref{fig:space} illustrates, reading from left to right, how we structure the crop-management decision as a directed acyclic graph. This representation builds upon the hierarchical design spaces framework~\cite{saves2026hierarchical,hallehannan2025distance} and relies on the \nolinkurl{adsg-core} implementation~\cite{bussemaker2024system}.
The two structural counts, the number of tillage passes and of fertilisations, are  \emph{meta variables} represented as \emph{choice nodes}, in blue, that fan out into the values they can take, or supports. In that case, their supports are $\{0,1,2\}$ and $\{0,\ldots,3\}$. Selecting a value for a meta variable \emph{decrees} some other \emph{decreed} variables, in yellow, whose domains are given under brackets such as dates and depths. For example, choosing 2 fertilisations activates the first and second date-dose pairs and leaves the third out. Three yellow variables, the sowing date and depth together with the harvest date, are \emph{neutral} and always active and independent of the other variables.
Two meta variables, the nine variables they decree, and three neutral ones make up the 14 parameters of the study. 
Consequently, the graph dictates which variables are active or inactive for any given itinerary. This process generates an \emph{activation vector} containing the values of only the active nodes, which we then carry through the analysis~\cite{bussemaker2025system}.
We model inactive variables as genuinely absent rather than imputing a numeric default. As such, a dose that is never applied is structurally deactivated. 
Regarding variables' types, \emph{continuous variables} are rescaled to a uniform range to prevent unit-based bias, while meta variables retain their integer encoding as \emph{ordinal variables}~\cite{saves2026hierarchical}.

\subsection{Learning What Matters with a Random Forest}
\label{ssec:rf}

A sensitivity analysis has to measure the impact of each parameter and of each combination of parameters, meaning computing $2^{14}-1$ terms, which can be time-consuming. With many weakly relevant inputs, the search for interactions quickly drowns in spurious combinations. We therefore first ask a simpler model which parameters are worth looking at. For each output, we fit a random forest~\cite{breiman2001random} and rank the parameters by \emph{permutation importance}, the intuition being that if shuffling a parameter barely changes the forest's predictions, then that parameter does not matter much for the output.
Two adjustments make this reading honest in a hierarchical space. The first concerns training. If inactive parameters were handed to the forest as a fixed default value, the forest could cheat, reading that recurring value as a tell-tale sign that a branch is switched off rather than learning the parameter's own effect. To remove that shortcut, we fill each inactive entry, before training, with a value drawn at random from the cases where the parameter is active, which scrambles the artificial signal while leaving the genuine relationships intact. The second concerns the importance measure itself. Shuffling a parameter across active and inactive rows alike would create itineraries that cannot exist, so we shuffle each parameter only among the itineraries where it is active, and we express the resulting loss of accuracy relative to how much the output varies within that active subspace of the graph. 
This \emph{conditional permutation importance} keeps a decision that applies rarely, deep in the design space graph, from being unfairly diluted by all the itineraries in which it plays no role.
These importances serve two purposes. They set how sharply each parameter is allowed to influence the dependence measure of the next step, and they let us discard parameters whose importance is indistinguishable from noise. In practice, a parameter whose importance falls below a small threshold ($\tau = 0.1\%$) is treated as noise and pruned, and the retained importances are scaled by a common factor ($s = 5.0$) that sets the length scales of the kernel of Section~\ref{ssec:hsic}. This value is large enough for a dominant parameter to be seen as almost unrelated to itself between the two ends of its range (a residual similarity of $\text{e}^{-5} \approx 0.7\%$), so that the most influential parameters are free to express their full effect.
 Ultimately, this pruning keeps the analysis affordable and sparse~\cite{Emine2025}. Traditional kernel tuning methods, like Gaussian process maximum likelihood estimation or kernel target alignment, struggle in such high-dimensional spaces; they spread the kernel weights too thinly across the many interactions, effectively getting lost in higher-order effects, which is a phenomenon referred to as the \emph{curse of dimensionality}~\cite{beh2025phase}.
\subsection{Measuring and Decomposing Dependence on the Design Space Graph}
\label{ssec:hsic}

To quantify how strongly a decision influences an output, we use the Hilbert-Schmidt Independence Criterion (HSIC)~\cite{gretton2005measuring,daveiga2015dependence}, a robust measure of statistical dependence. Unlike standard correlation, HSIC detects nonlinear and threshold-like relationships, which is essential for capturing agronomic responses, such as yield saturating once nitrogen needs are met. HSIC compares configurations through a kernel that acts as a similarity measure between input parameters.

The validity of this approach rests entirely on making that similarity respect the hierarchical structure of the parameter space. To achieve this, we apply the conditional algebraic distance~\cite{hallehannan2025distance,saves2024smt}. For an operation that an upstream choice can switch on or off, two itineraries that both omit it are treated as identical on that operation, since there is nothing to compare. One that operates and one that omits it are treated as maximally different, as this gap represents a discrete decision. Finally, two itineraries that both perform it are compared via an ordinary numeric distance on their continuous parameters (\textit{e.g.}, date, dose, or depth). Structural counts, such as the number of tillage passes or fertilisations, are handled separately, because a count selects which sub-itinerary is instantiated rather than tuning a quantity~\cite{saves2026hierarchical,halstrup2016black}. 
Moving from one to two fertilisations is a change in management structure, not a mere numerical step. Consequently, switched-off branches contribute no artificial variability, ensuring the measured dependence reflects actual decisions rather than imputed placeholder values.

In practice, the empirical HSIC reduces to the Frobenius inner product of the inputs' and outputs' centred Gram matrices~\cite{gretton2005measuring,daveiga2015dependence,deLozzo2016hsic,spagnol2019hsic}. While this marginal formulation is highly effective for screening complex simulators~\cite{marrel2021statistical}, it evaluates variables in isolation~\cite{jaber2026fusion}. To resolve this, we apply a kernel-based functional ANOVA~\cite{durrande2013anova,daveiga2021anova,novello2022making} to split the total input-output dependence into orthogonal shares. This orthogonality is guaranteed provided the univariate kernels are empirically centred, a condition our pipeline satisfies by construction. Each variable receives a main-effect share, and each combination receives an interaction share, all expressed as fractions of the total dependence. Because the random forest has already pruned irrelevant parameters, the remaining combinatorial space is small enough to evaluate interactions of up to four simultaneous decisions. This yields the ranked decompositions presented in Section~\ref{sec:results}, achieved with a highly tractable computational complexity of $\mathcal{O}(n^2 p)$, where $n$ is the number of simulations and $p$ the number of explored variables.

\subsection{Comparing Rare and Common Decisions Fairly}
\label{ssec:intrinsic}
Even with normalisation, one bias remains that is specific to graph-structured spaces. A decision that is only active in a small fraction of the itineraries can never top a global ranking simply because it is absent from most of the samples. The importance of a third fertilisation, present in only a quarter of the itineraries, is structurally reduced.
As HSIC compares itineraries pairwise, and because a pair of itineraries says something about a group of decisions only when that group is present in both itineraries, we propose to shift from ``how much does this decision move the output across all itineraries?'' to ``how much does it move the output among the itineraries where it is actually taken?''. 
The hierarchical distance confines the signal of a subset $A$ of decisions to the itineraries where $A$ is jointly active, that is, with probability $p_A$. Then, only about $p_A^2$ of all pairs are informative and
$
    S^{\text{intrinsic}}_{A} = \frac{\widehat{\text{HSIC}}_A}{p_A^{2}}
$
recovers the influence that a decision carries when it is taken. A following renormalisation step ensures that the total shares sum up to $100\%$ to give a new \emph{intrinsic index}. This intrinsic reading places deeply nested choices on the same footing as the always-active meta variables and gives a complementary view of the model's levers, alongside global ranking.

\subsection{Novelty with Respect to the State of the Art}
\label{ssec:novelty}

Each step above draws on established ideas, but their combination joins two lines of work that had stayed apart. On one side, kernel-based dependence measures and their ANOVA decompositions~\cite{daveiga2015dependence,daveiga2021anova} rank nonlinear and interacting effects efficiently, yet, as applied to global sensitivity analysis, they assume inputs that vary freely and independently and provide no built-in handling of variables that exist only conditionally, so applying them here forces the analyst to invent values for absent variables and biases the result. On the other side, the machinery for hierarchical design spaces and its conditional distances~\cite{saves2026hierarchical,bussemaker2025system,hallehannan2025distance} handles exactly such trees, but has so far served optimisation and prediction, not GSA. 

While recent efforts have explored defining Sobol' indices for hierarchical variables~\cite{bartok2023defining}, establishing exact functional ANOVA decompositions for categorical inputs~\cite{ferrere2026exact}, and quantifying the contribution of variable subsets in the presence of dependencies~\cite{labreuche2024contribution}, they still do not fully resolve the challenge of conditionally active variables within a unified global sensitivity analysis. 
Our pipeline addresses this gap by bringing the conditional distances into the dependence measure.  By relying on this formalisation, we use the random forest to keep the interaction analysis tractable, and add the activation-frequency correction so that conditionally active decisions can finally be ranked within a single, structure-aware decomposition. To make this reusable beyond the present study, hierarchical structures and samplers are based on the extension \nolinkurl{smt-design-space-ext} of SMT~\cite{saves2026hierarchical}. This open project is reproducible online\footnote{See \url{https://anonymous.4open.science/r/Sensitivity_analysis_MAELIA-DB70}}.

\section{Results}
\label{sec:results}

\subsection{Experimental Setup}
\label{sec:setup}

\noindent {\textbf{Study configuration and outputs.}}
The pipeline of Section~\ref{sec:method} is applied to the homogenised \emph{terrainSA} configuration, in which a single reference plot is cloned so that soil type, geometry and weather zone are held constant across simulations. This deliberate homogenisation removes the dominant pedo-climatic signal and isolates the effect of the cropping itinerary, which is the object of the sensitivity analysis. For the same reason, the analysis is restricted to a single crop, winter wheat; focusing on one well-characterised production system removes the confounding influence of crop identity and rotation effects, so that the observed output variability can be attributed to the technical operations themselves rather than to differences between crops. Three agronomic and environmental outputs are analysed, each aggregated over the simulated horizon: $\mathrm{N_{lixi}}$, $\Delta\mathrm{C_{org}}$, and yield.

\noindent {\textbf{Design of experiments.}}
The design comprises $5,000$ configurations drawn from the hierarchical space of Fig.~\ref{fig:space} by hierarchical Latin hypercube sampling with the \texttt{smt-design-space-ext} extension of SMT, so that the conditional activation of decreed variables is respected at sampling time~\cite{saves2024smt,saves2026hierarchical}. Each configuration is simulated over the ten weather years of the record. Its outputs are computed at the end of each year and averaged, which damps the interannual variability of the weather.
 Continuous variables are rescaled to $[0,1]$ on their observed bounds, the two structural counts keep their integer codes, and inactive variables carry a dedicated inactive state for identification as described in Section~\ref{ssec:encoding}.

\noindent {\textbf{Analysis settings.}}
For each output, the impact of the inputs is evaluated over the full set of $5,000$ simulations up to interaction of order 4. To do this, we leverage random forests to supervise kernel scales through the conditional permutation importance of Section~\ref{ssec:rf}. Then, the HSIC-ANOVA decomposition of Section~\ref{ssec:hsic} builds a Gaussian input kernel from the conditional algebraic distance on the conditionally active continuous variables and an ordinal distance on the two structural counts, together with a Gaussian output kernel whose bandwidth follows the output standard deviation. A standard heuristic and a fixed random seed make the forest and the permutations reproducible; the terms contributing below a small fraction of the total dependence are discarded.  The random forest uses $300$ unconstrained trees with a minimum leaf size of $1\%$ of the sample; inactive entries are filled by random marginal imputation, and the conditional permutation importance is averaged over $6$ reshuffles per variable. 
\subsection{Per-Output Sensitivity}
\label{ssec:per-output}

Figure~\ref{fig:hsic} reports, for each output, the HSIC-ANOVA terms ranked by their share of the global input-output dependence, a first-order term being a single-variable main effect and a higher-order term an interaction. Across the three outputs, the bulk of the dependence is carried by first-order effects, and the few interactions above the retention threshold almost all couple the number of fertilisations with the applied doses. The itinerary therefore acts through additive main effects, with a secondary interaction structure concentrated on the fertilisations.

For nitrogen leaching, the leading levers are the fertiliser applications, through both their doses and their dates: the second dose (\nolinkurl{Second\_Fertilisation\_Dose}, $12.1\%$), the number of fertilisations (\nolinkurl{Number\_Of\_Fertilisations}, $9.4\%$), the second date ($9.2\%$), the first dose ($8.6\%$) and the third dose ($6.9\%$). This is consistent with an output driven by the synchronisation between nitrogen supply and the crop uptake window, in which mistimed or oversized applications leave mineral nitrogen available for deep drainage.


\begin{figure}[tb!]
\centering

\definecolor{colorOrder1}{rgb}{0.01, 0.47, 0.51}
\definecolor{colorOrder2}{rgb}{0.89, 0.67, 0.18}
\definecolor{colorOrder3}{rgb}{0.85, 0.47, 0.26}

\resizebox{\textwidth}{5.2cm}{%
\begin{tikzpicture}[font=\sffamily]

\begin{groupplot}[
    group style={
        group size=3 by 1,
        horizontal sep=3.75cm, 
        xlabels at=edge bottom,
        xticklabels at=edge bottom,
    },
    height=6cm, 
    width=6.5cm,
    xmin=0, xmax=26,
    ymin=0.2, ymax=10.5, 
    xtick={0,5,10,15,20,25},
    ytick={1,2,3,4,5,6,7,8,9,10}, 
    yticklabel style={font=\tiny, align=right, anchor=east, text width=4.2cm, inner sep=1.9pt},
    xlabel={Contribution to global HSIC (\%)},
    xlabel style={font=\footnotesize},
    title style={font=\footnotesize\bfseries, yshift=1ex},
    point meta=x, 
    nodes near coords={\pgfmathprintnumber[fixed, precision=1, fixed zerofill]{\pgfplotspointmeta}\%},
    nodes near coords align={horizontal},
    nodes near coords style={font=\tiny, text=black},
    grid=major,
    grid style={line width=.2pt, draw=gray!30},
    axis x line=bottom,
    axis y line=left,
    axis line style={draw=none},
    x tick label style={font=\tiny}
]

\nextgroupplot[
    title={Nitrogen leaching},
     xmin=0, xmax=15,
    xtick={0,5,10,15},
    yticklabels={
        \textbf{1st Ferti. Date x 1st Ferti. Dose},
        \textbf{Nb. Prepa.},
        \textbf{Nb. Ferti. x 3rd Ferti. Dose},
        \textbf{Nb. Ferti. x 1st Ferti. Dose},
        \textbf{1st Ferti. Date},
        \textbf{3rd Ferti. Dose},
        \textbf{1st Ferti. Dose},
        \textbf{2nd Ferti. Date},
        \textbf{Nb. Ferti.},
        \textbf{2nd Ferti. Dose},
    }
]

\addplot[color=colorOrder1, xcomb, mark=none, line width=8.75pt] coordinates {(12.1,10) (9.4,9) (9.2,8) (8.6,7) (6.9,6) (3.6,5) (2.5,2) };
\addplot[color=colorOrder2, xcomb, mark=none, line width=8.75pt] coordinates {(3.4,4) (2.6,3) (2.0,1)};

\nextgroupplot[
    title={Soil organic carbon change},
    xmin =0, xmax=27.5,
    xtick={0,5,10,15,20,25},
    yticklabels={
        \textbf{Nb. Ferti. x 1st Ferti. Dose x 2nd Ferti. Dose},
        \textbf{Nb. Ferti. x Harvest Date x 2nd Ferti. Dose},
        \textbf{Nb. Ferti. x Harvest Date x 1st Ferti. Dose},
        \textbf{3rd Ferti. Dose},
        \textbf{Nb. Ferti. x 1st Ferti. Dose},
        \textbf{Nb. Prepa.},
         \textbf{Nb. Ferti.},
         \textbf{1st Ferti. Dose},
         \textbf{2nd Ferti. Dose},
         \textbf{Harvest Date}
    }
]
\addplot[color=colorOrder1, xcomb, mark=none, line width=8.75pt] coordinates {(25.8,10) (12.8,9) (11.2,8) (10.2,7) (8.5,6) (3.0,4)};
\addplot[color=colorOrder2, xcomb, mark=none, line width=8.75pt] coordinates {(3.4,5)};
\addplot[color=colorOrder3, xcomb, mark=none, line width=8.75pt] coordinates {(1.8,3) (1.6,2) (1.5,1)};

\nextgroupplot[
    title={Yield},
    xmin =0, xmax=25,
    xtick={0,5,10,15,20,25},
    yticklabels={
        \textbf{1st Ferti. Dose x 2nd Ferti. Dose},
        \textbf{Nb. Ferti. x 2nd Ferti. Dose x 3rd Ferti. Dose},
        \textbf{Nb. Ferti. x 2nd Ferti. Dose},
        \textbf{Nb. Ferti. x 3rd Ferti. Dose},
        \textbf{Nb. Ferti. x 1st Ferti. Dose x 2nd Ferti. Dose},
        \textbf{Nb. Ferti. x 1st Ferti. Dose},
        \textbf{3rd Ferti. Dose},
        \textbf{Nb. Ferti.},
        \textbf{1st Ferti. Dose},
        \textbf{2nd Ferti. Dose},
    }
]
\addplot[color=colorOrder1, xcomb, mark=none, line width=8.75pt] coordinates {(20.6,10) (19.0,9) (17.6,8) (8.7,7)};
\addplot[color=colorOrder2, xcomb, mark=none, line width=8.75pt] coordinates {(7.7,6) (3.8,4) (2.5,3) (2.3,1)};
\addplot[color=colorOrder3, xcomb, mark=none, line width=8.75pt] coordinates {(4.7,5) (2.4,2)};

\end{groupplot}

\path (group c2r1.south) +(0,-1.4cm) node[anchor=north] (legend) {
    \begin{tikzpicture}[font=\footnotesize\sffamily, x=3cm, y=10pt]
        \fill[colorOrder1] (0,0.1) rectangle (0.4, 0.8) node[right, pos=1.0, black] {Order 1};
        \fill[colorOrder2] (1,0.1) rectangle (1.4, 0.8) node[right, pos=1.0, black] {Order 2};
        \fill[colorOrder3] (2,0.1) rectangle (2.4, 0.8) node[right, pos=1.0, black] {Order 3};
    \end{tikzpicture}
};

\end{tikzpicture}%
}

\caption{HSIC-ANOVA top terms for the three outputs, ranked by contribution to the global input-output dependence. Most retained terms are first-order main effects, with the strongest interactions coupling the number of fertilisations with the applied doses.}
\label{fig:hsic}
\end{figure}
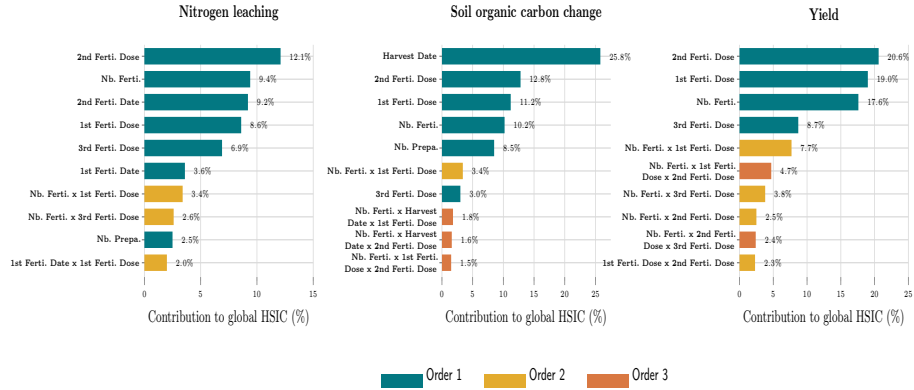

Soil organic-carbon variation is governed by a different family. The harvest date (\nolinkurl{Harvest\_Date}, $25.8\%$) carries by far the largest single share, ahead of the second and first doses ($12.8\%$ and $11.2\%$) and the numbers of fertilisations ($10.2\%$) and of preparations ($8.5\%$). The length of the crop cycle sets the duration over which residues are returned, and organic matter is mineralised, while the coarse structural choices of the itinerary modulate the buried carbon and its turnover. It is also the output for which higher-order interactions are most visible, with three third-order terms above the threshold, two coupling the number of fertilisations with the harvest date and an early dose and the third joining it to the first two doses, although each stays small.

Yield is the output most clearly driven by the quantity of nitrogen supplied. The second, first and third doses ($20.6\%$, $19.0\%$ and $8.7\%$) together with the number of fertilisations ($17.6\%$) dominate its profile, and it carries the strongest interaction structure of the three, coupling the number of fertilisations with the first dose ($7.7\%$) and, at third order, with the first two doses ($4.7\%$).

Note that the meta-variables, starting with the number of fertilisations, rank among the most important parameters, showing that the coarse composition of the itinerary matters more than the fine tuning of its operations. Also, the shares plotted in Fig.~\ref{fig:hsic} are \emph{global}, measured over the whole design, so a decreed variable active on only part of the itineraries is mechanically penalised: the third fertiliser application, active on about $25\%$ of the design against $50\%$ for the second and $75\%$ for the first, comes last of the three doses in every output.
\begin{table}[tb!]
\centering
\caption{Global versus intrinsic (activation-adjusted by $1/p_A^2$) HSIC share of the three fertiliser-dose main effects, expressed in global dependence shares.} 
\label{tab:intrinsic}
\setlength{\tabcolsep}{6pt}
\resizebox{\linewidth}{1.1cm}{
\begin{tabular}{@{}l cc cc cc@{}}
\toprule
 & \multicolumn{2}{c}{Nitrogen leaching} & \multicolumn{2}{c}{SOC change} & \multicolumn{2}{c}{Yield} \\
\cmidrule(lr){2-3}\cmidrule(lr){4-5}\cmidrule(lr){6-7}
Fertiliser dose & global & intrinsic & global & intrinsic & global & intrinsic \\
\midrule
First ($p_A = 0.75$)  & $8.6$  & $2.6$  & $11.2$ & $6.9$  & $19.0$ & $6.6$  \\
Second ($p_A = 0.50$) & $12.1$ & $8.2$  & $12.8$ & $17.8$ & $20.6$ & $16.1$ \\
Third ($p_A = 0.25$)  & $6.9$  & $19.3$ & $3.0$  & $17.5$ & $8.7$  & $28.0$ \\
\bottomrule
\end{tabular}
}
\end{table}
The intrinsic reading of Section~\ref{ssec:intrinsic} removes this bias to show how much a decision moves the output \emph{among the itineraries where it is actually applied}. For yield, the two readings disagree sharply as depicted in Table~\ref{tab:intrinsic}: the global ranking of the doses is second ($20.6\%$), first ($19.0\%$), third ($8.7\%$), whereas the intrinsic one reverses it to third ($28.0\%$), second ($16.1\%$), first ($6.6\%$). The third dose seems the most decisive dose when a farmer chooses to make 3 fertilisations. The same reversal holds for nitrogen leaching, and the late dose climbs to a near-first position for soil organic carbon. These two readings respond to different questions: the global one weighs a decision by how much it shapes the outputs over the whole population of itineraries, while the intrinsic one by how much it matters when it is on the table.
No variable pruned by the random forest reappears in the HSIC-ANOVA table of its output, which confirms that the supervision of Section~\ref{ssec:rf} concentrates the decomposition on the actionable levers. 

\subsection{Aggregation by Decision Family}
\label{ssec:groups}
To read the results at the level of management decisions rather than individual parameters, the per-term contributions are aggregated into four semantic families: \emph{fertilisation doses}, \emph{fertilisation timing}, the \emph{sowing and harvest calendar} (also carrying preparation-to-sowing delay), and the structural itinerary composition, that is the number of fertilisations and of preparations, which we call \emph{management structure}. Because an interaction term belongs to several families, its contribution is divided equally among its constituent parameters before aggregation, so that family shares remain additive. A fifth family, the working depths of tillage and sowing, carries negligible dependence for all three outputs and is omitted. 

Figure~\ref{fig:hsic-groups} reports, for each output, the fractional share of the global dependence carried by each family.
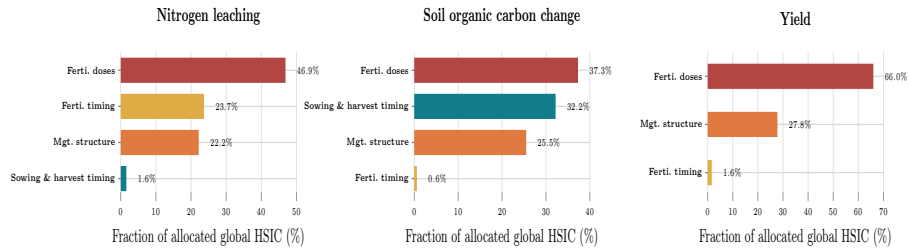
\begin{figure}[b!]
\centering

\definecolor{colorRank1}{rgb}{0.70, 0.27, 0.26} 
\definecolor{colorRank2}{rgb}{0.88, 0.48, 0.25} 
\definecolor{colorRank3}{rgb}{0.87, 0.67, 0.27} 
\definecolor{colorRank4}{rgb}{0.07, 0.49, 0.54} 

\resizebox{\textwidth}{3.33cm}{%
\begin{tikzpicture}[font=\sffamily]

\begin{groupplot}[
    group style={
        group size=3 by 1,
        horizontal sep=3.3cm, 
        xlabels at=edge bottom,
        xticklabels at=edge bottom,
    },
    height=4cm, 
    width=6.5cm,
    xmin=0, 
    yticklabel style={font=\tiny, align=right, anchor=east, text width=3.5cm, inner sep=2pt},
    xlabel={Fraction of allocated global HSIC (\%)},
    xlabel style={font=\footnotesize},
    title style={font=\footnotesize\bfseries, yshift=1ex},
    point meta=x, 
    nodes near coords={\pgfmathprintnumber[fixed, precision=1, fixed zerofill]{\pgfplotspointmeta}\%},
    nodes near coords align={horizontal},
    nodes near coords style={font=\tiny, text=black},
    grid=major,
    grid style={line width=.2pt, draw=gray!30},
    axis x line=bottom,
    axis y line=left,
    axis line style={draw=none},
    x tick label style={font=\tiny}
]

\nextgroupplot[
    title={Nitrogen leaching},
    xmax=50, xtick={0,10,20,30,40,50}, 
    ymin=0.5, ymax=4.5, ytick={1,2,3,4},
    yticklabels={
        \textbf{Sowing \& harvest timing},
        \textbf{Mgt. structure},
        \textbf{Ferti. timing},
        \textbf{Ferti. doses}
    }
]
\addplot[color=colorRank1, xcomb, mark=none, line width=12pt] coordinates {(46.9,4)};
\addplot[color=colorRank3, xcomb, mark=none, line width=12pt] coordinates {(23.7,3)};
\addplot[color=colorRank2, xcomb, mark=none, line width=12pt] coordinates {(22.2,2)};
\addplot[color=colorRank4, xcomb, mark=none, line width=12pt] coordinates {(1.6,1)};

\nextgroupplot[
    title={Soil organic carbon change},
    xmax=40, xtick={0,10,20,30,40},
    ymin=0.5, ymax=4.5, ytick={1,2,3,4},
    yticklabels={
        \textbf{Ferti. timing},
        \textbf{Mgt. structure},
        \textbf{Sowing \& harvest timing},
        \textbf{Ferti. doses}
    }
]
\addplot[color=colorRank1, xcomb, mark=none, line width=12pt] coordinates {(37.3,4)};
\addplot[color=colorRank4, xcomb, mark=none, line width=12pt] coordinates {(32.2,3)};
\addplot[color=colorRank2, xcomb, mark=none, line width=12pt] coordinates {(25.5,2)};
\addplot[color=colorRank3, xcomb, mark=none, line width=12pt] coordinates {(0.6,1)};

\nextgroupplot[
    title={Yield},
    xmax=70, xtick={0,10,20,30,40,50,60,70},
    ymin=0.5, ymax=3.5, ytick={1,2,3}, 
    yticklabels={
        \textbf{Ferti. timing},
        \textbf{Mgt. structure},
        \textbf{Ferti. doses}
    }
]
\addplot[color=colorRank1, xcomb, mark=none, line width=12pt] coordinates {(66.0,3)};
\addplot[color=colorRank2, xcomb, mark=none, line width=12pt] coordinates {(27.8,2)};
\addplot[color=colorRank3, xcomb, mark=none, line width=12pt] coordinates {(1.6,1)};

\end{groupplot}

\end{tikzpicture}%
}
\caption{HSIC-ANOVA contributions aggregated into four decision families. The sowing and harvest calendar carries no sufficient and retained term for yield.}
\label{fig:hsic-groups}
\end{figure}
This aggregation contrasts outputs explicitly. Fertilisation doses are the most influential family for all three outputs, but the family that comes next differs from one output to the other. Nitrogen leaching is a fertilisation matter above all, with doses ($46.9\%$) and timing ($23.7\%$) together accounting for about 70\% of its dependence, the management structure adding a further $22.2\%$ and the sowing and harvest calendar staying marginal ($1.6\%$). Soil organic-carbon variation spreads its dependence more evenly across doses ($37.3\%$), the sowing and harvest calendar ($32.2\%$) and management structure ($25.5\%$), fertilisation timing being negligible ($0.6\%$), reflecting an output shaped by nitrogen quantities, how long the crop occupies the ground and the itinerary composition rather than by the exact application dates. Yield is the most concentrated output, with fertilisation doses carrying about two-thirds of its dependence alone ($66.0\%$) ahead of management structure ($27.8\%$) while the timing family contributes little ($1.6\%$).
Together, these panels show that a single hierarchical sensitivity analysis separates the outputs by their most influential levers. While fertilisation doses lead all outputs, the discriminating secondary family is fertilisation timing for leaching, the crop calendar for carbon, and the management structure for yield. The latter is a substantial contributor to all three outputs. 

\section{Conclusion and Perspectives}
\label{sec:conclusion}

This paper argued that a complex agro-ecological agent-based model should not be explored as if its parameters formed an unstructured hypercube, because its agricultural decisions are intrinsically hierarchical: many parameters exist only once a higher-level management choice activates them. Taking the MAELIA platform as a case study, the crop-management itinerary, expressed through date-dose formalism, was encoded as a hierarchical design space of meta and decreed variables, and a random forest supervised HSIC-ANOVA was defined over it. Applied to three outputs on a homogenised, single-crop configuration, the analysis returned an interpretable, decision-oriented picture. Fertilisation doses are the leading lever for all three outputs, and the family that comes next sets them apart: fertilisation timing for nitrogen leaching, the sowing and harvest calendar for soil organic-carbon variation, and the structural composition of the itinerary for yield, with the number of fertilisations standing out as a top-ranked decision throughout. First-order effects carry most of the dependence, and the interactions that emerge, mostly couplings between the number of fertilisations and the applied doses, stay secondary. These conclusions follow from an analysis carried out within the structured space in which the decisions are actually made.

A natural question is how this pipeline compares against established sensitivity-analysis methods such as variance-based GSA indices~\cite{IlIdrissi2025BeyondShapley}. 
In our setting, however, no such head-to-head comparison is well posed. Classical variance-based methods presuppose inputs that vary independently over a fixed Cartesian product, an assumption that a branched design space violates by construction: a large fraction of the combinations they would require assign values to sub-parameters whose activating branch is switched off, and are therefore configurations the simulator never produces. A Sobol' analysis run over the flattened space would thus not be a competing measurement of the same quantity, but an answer to a different, ill-posed question in which the conditional structure has been discarded. More generally, forcing a classical method onto this space requires imputing arbitrary values for the inactive variables, and these imputed values distort the variance decomposition, biasing the resulting indices and masking the influence of the upstream decisions that control each branch. The contribution here is to define a sensitivity attribution that remains well-posed on the hierarchical space, not to outperform a baseline that does not address the same problem~\cite{idrissi2025hoeffding}.

Several directions extend this work. On the modelling side, reintroducing controlled pedo-climatic variation and additional crops would test how robust the decision-family rankings are once the environment and the crop are allowed to vary again, and applying the same pipeline to other GAMA-based agro-ecological models would probe its generality. Finally, because the encoding and the sampler are shared with surrogate-based optimisation, the same structured space could support Bayesian optimisation of management itineraries, turning the sensitivity analysis into a first step of a broader decision-support workflow~\cite{saves2026hierarchical2}. 
Furthermore, on the visualisation side, and to enhance the interpretability of our results, we developed an interactive web page featuring post-hoc explainable AI tools such as partial dependence plots and individual conditional expectation curves. This interface leverages the \nolinkurl{smt-explainability} library~\cite{robani2025smt} to allow for real-time visual exploration of the agronomic responses for stakeholders and thematicians.

\begin{credits}
\textbf{\ackname}
This work has received funding managed by the Agence Nationale de la Recherche under the France 2030 program for the SLAM-B project (ANR-22-PEXF-0003) part of the PEPR FairCarboN and the MIMICO project (ANR-24-CE23-0380).

\end{credits}

\clearpage
\newpage
\noindent 
\begin{credits}
\noindent \hspace{-0.2cm}\textbf{\discintname}
The authors have no competing interests to declare that are relevant to the content of this article. \\
%
%
\noindent\textbf{AI Declaration}
The authors declare that generative AI was used in writing this manuscript. 
\end{credits}
\
\bibliographystyle{splncs04}

\bibliography{main}

\end{document}